\documentclass[sigconf]{aamas}

\usepackage{balance} 
\usepackage{bm}
\usepackage{graphicx}
\usepackage{subcaption}
\usepackage{multirow}

\usepackage{textgreek}
\usepackage{array}
\usepackage{booktabs}
\usepackage[inline]{enumitem}
\usepackage{tabularx}
\usepackage{graphicx}
\makeatletter
\gdef\@copyrightpermission{
  \begin{minipage}{0.2\columnwidth}
   \href{https://creativecommons.org/licenses/by/4.0/}{\includegraphics[width=0.90\textwidth]{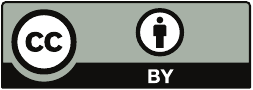}}
  \end{minipage}\hfill
  \begin{minipage}{0.8\columnwidth}
   \href{https://creativecommons.org/licenses/by/4.0/}{This work is licensed under a Creative Commons Attribution International 4.0 License.}
  \end{minipage}
  \vspace{5pt}
}
\makeatother

\setcopyright{ifaamas}
\acmConference[AAMAS '25]{Proc.\@ of the 24th International Conference
on Autonomous Agents and Multiagent Systems (AAMAS 2025)}{May 19 -- 23, 2025}
{Detroit, Michigan, USA}{Y.~Vorobeychik, S.~Das, A.~Nowé  (eds.)}
\copyrightyear{2025}
\acmYear{2025}
\acmDOI{}
\acmPrice{}
\acmISBN{}

\acmSubmissionID{<<OpenReview submission id>>}

\title{Multi-Objective Reinforcement Learning for Water Management}

\subtitle{Extended Abstract}

\author{Zuzanna Osika}
\affiliation{
  \institution{Delft University of Technology}
  \city{Delft}
  \country{Netherlands}}
\email{z.osika@tudelft.nl}

\author{Roxana R\u{a}dulescu}
\affiliation{
  \institution{Utrecht University}
  \city{Utrecht}
  \country{Netherlands}}
\email{r.t.radulescu@uu.nl}

\author{Jazmin Zatarain-Salazar}
\affiliation{
  \institution{Delft University of Technology}
  \city{Delft}
  \country{Netherlands}}
\email{j.zatarain-salazar@tudelft.nl}

\author{Frans A. Oliehoek}
\affiliation{
  \institution{Delft University of Technology}
  \city{Delft}
  \country{Netherlands}}
\email{f.a.oliehoek@tudelft.nl}

\author{Pradeep K. Murukannaiah}
\affiliation{
  \institution{Delft University of Technology}
  \city{Delft}
  \country{Netherlands}}
\email{p.k.murukannaiah@tudelft.nl}

\begin{abstract}
Many real-world problems (e.g., resource management, autonomous driving, drug discovery) require optimizing multiple, conflicting objectives. Multi-objective reinforcement learning (MORL) extends classic reinforcement learning to handle multiple objectives simultaneously, yielding a set of policies that capture various trade-offs. However, the MORL field lacks complex, realistic environments and benchmarks. We introduce a water resource (Nile river basin) management case study and model it as a MORL environment. We then benchmark existing MORL algorithms on this task. Our results show that specialized water management methods outperform state-of-the-art MORL approaches, underscoring the scalability challenges MORL algorithms face in real-world scenarios.
\end{abstract}



\newcommand{\BibTeX}{\rm B\kern-.05em{\sc i\kern-.025em b}\kern-.08em\TeX}

\begin{document}
\emergencystretch 3em

\pagestyle{fancy}
\fancyhead{}


\maketitle 


\section{Introduction}

Multi-objective reinforcement learning (MORL) extends the classic reinforcement learning (RL) framework \cite{Sutton1998} to problems with multiple, conflicting objectives. Rather than maximizing a single reward, MORL uses a vector of rewards---one per objective---yielding multiple optimal policies with distinct trade-offs \cite{hayes2022practical}. It has been applied in water management \cite{casteletti-momd-2012,giuliani2016curses}, autonomous driving \cite{li-urban-2019}, power allocation \cite{oh2023multi,xiong2023multi}, drone navigation \cite{wu2024multi}, and medical treatment \cite{jalalimanesh2017multi,lizotte-2010-efficient}. Recent advances \cite{alegre2023gpi,felten_toolkit_2023,xu2020prediction} have expanded MORL's algorithmic capabilities, yet benchmarking still heavily relies  on toy tasks. As a result, the real-world applicability of state-of-the-art algorithms remains unclear, especially given the additional objective dimension alongside large state and action spaces.

Water resource management is becoming increasingly complex due to climate change. A key problem in this domain is about making sequential decisions about water releases from dams to balance multiple, often conflicting objectives, including hydropower generation, irrigation, water supply, and environmental preservation. This multi-objective problem has been studied in real-world contexts \cite{casteletti-momd-2012,giuliani2014many,zaniolo2021neuro,owusu2022quantifying} using methods such as dynamic programming and mathematical optimization \cite{herman2020climate,giuliani2021state}. A specialized method, evolutionary multi-objective direct policy search (EMODPS) \cite{giuliani2016curses}, has demonstrated strong performance \cite{quinn2019controlling,salazar2016diagnostic}. However, existing models of the problem remain difficult to access due to proprietary data, multiple programming languages, and inconsistent code structures.

In this paper, we address the problem of the lack of large-scale, real-world problems for evaluating MORL algorithms. We introduce a real-world case study from the field of water resource management and model it in a modular, structured way as a MORL environment using MO-Gymnasium \cite{felten_toolkit_2023} API. This case study enables MORL researchers to test their algorithms in realistic, high-impact scenarios. The case study we introduce is Nile river basin management. By benchmarking state-of-the-art MORL algorithms in this environment, we evaluate their performance against the specialized water management algorithm, EMODPS, and explore the effectiveness of existing MORL methods for real-world application.\footnote{Code: \url{https://github.com/osikazuzanna/wms_morl/}}

\section{Background}

We formalize a MORL problem as a multi-objective Markov decision process (MOMDP) 
\(\langle S, A, T, \gamma, \mu, \mathbf{R} \rangle\), where \(S\) and \(A\) are the state and action spaces, 
\(T\) is a probabilistic transition function, \(\gamma\) is a discount factor, \(\mu\) is the initial state distribution, 
and \(\mathbf{R}\) is a vector-valued reward function returning a numeric feedback signal for each of \(d \geq 2\) objectives—the key difference from standard MDPs. 

Because the policy value \(\mathbf{V}^\pi \in \mathbb{R}^d\) can only be partially ordered, either a utility function 
\(u: \mathbb{R}^d \to \mathbb{R}\) is used to prioritize objectives or, if unknown, an entire solution set (multi-policy paradigm) must be returned.
The \textbf{Pareto set} is the set of all pairwise undominated policies based on Pareto dominance, while the \textbf{Pareto front} (PF) is the corresponding set of value vectors, forming the \textbf{convex hull} (CH) if \(u\) is a positively weighted linear sum \cite{Osika2024}.

\section{Nile River Basin}

\begin{description}[leftmargin=0em,itemsep=0.5em]
\item [Background]
The Nile River, a vital resource for ten northeast African countries, supports hydropower, agriculture, and municipal needs but faces ongoing political tensions over water rights, particularly between Egypt, Sudan, and Ethiopia. Ethiopia's Grand Ethiopian Renaissance Dam (GERD), seen as essential for economic development and expanding electricity access (currently only 27\%), is a major point of contention. Egypt, historically reliant on the Nile and the High Aswan Dam (HAD) for hydropower and water security, fears GERD will reduce downstream flows, jeopardizing food and energy security. Sudan, benefiting from extensive irrigation infrastructure, initially opposed GERD alongside Egypt but now acknowledges potential flood regulation benefits while still expressing concerns about water security. Negotiations remain stalled as the river faces additional challenges, including variable inflows, droughts, floods, and increasing demand due to population growth \cite{wheeler2018exploring}. Our study models the Nile River Basin to capture the conflicting interests of Ethiopia, Sudan, and Egypt.

\item[Actions] The action tuple is four-dimensional, where each dimension
represents the percentage of water to be released from one of
the following dams: GERD, Roseires, Sennar, and HAD. The actions are taken every month, for 20 years. Thus, the full episode consists of 240 timesteps.

\item[Observations]
The state is five-dimensional, with the components representing the storage of the four reservoirs modeled and the month of the year the system is in (normalized).

\item[Rewards] We consider four objectives; so, the reward is four-dimensional: Egypt irrigation deficit (ED), Sudan irrigation deficit (SD), Egypt minimum HAD level, which represents water reliability in the HAD ensuring that the water level exceeds the minimum required for power generation (HAD), and Ethiopia hydropower generation from the GERD power plant (EH).
\end{description}

\section{Experiments and Results}
\label{subsec:morl_alg}

\begin{description}[leftmargin=0em,itemsep=0.5em]
\item[Experimental Setup] We benchmark three state-of-the-art multi-policy MORL algorithms, covering both CH and PF-returning methods: GPI-LS \citep{alegre2023gpi}, PCN \cite{reymond_pareto_2022} and CAPQL \cite{Lu-2023-ICLR-CAPQL}. These algorithms were chosen specifically as they can handle continuous actions and state space as well as more than two objectives. These are compared with EMODPS \citep{giuliani2016curses}, an
advanced optimization approach for designing efficient control policies in complex water management systems. To compare the performance of these algorithms we use hypervolume (region between the solution set and a reference point, which defines the lower bound for each objective) ($\uparrow$), cardinality (number of solutions in the final set) (\rotatebox[origin=c]{-45}{$\uparrow$}) and sparsity (spread of the solutions in the solution set) ($\downarrow$) of the solution set \cite{xu2020prediction}. We trained each MORL algorithm for 200,000 timesteps and EMODPS for 20,000 NFEs, selecting these values based on observed convergence.

\item[Results]
Table~\ref{tab:results} shows final mean metrics (over five seeds) and uses EMODPS hypervolume as the baseline (value=1 indicates matching performance). GPI-LS achieves the highest MORL hypervolume (68\% of the baseline), while PCN and CapQL remain at 10–11\%. EMODPS outperforms in cardinality, which measures the number of solutions in the final set. Despite fewer solutions and higher sparsity, GPI-LS covers diverse trade-offs, whereas PCN and CapQL concentrate on narrower solution regions.

\end{description}

\begin{table}[!htb]
    \centering
    \begin{tabular}{@{}l@{\hspace{.1cm}}r@{\hspace{.1cm}}r@{\hspace{.2cm}}r@{}}
    \toprule
    \textbf{Algorithm} & \multicolumn{1}{p{1.9cm}}{\centering \textbf{Hypervolume\\(\% Baseline) $\uparrow$}} & \multicolumn{1}{p{1.4cm}}{\centering \textbf{Cardinality \rotatebox[origin=c]{-45}{$\uparrow$}}} & \multicolumn{1}{p{.9cm}}{\centering \textbf{Sparsity $\downarrow$}} \\
    \midrule
    EMODPS (Baseline) & 2.21E+08 (100\%) & 327 & 3.951 \\
    PCN & 2.26E+07 \phantom{0}(10\%) & 51 & 7.754 \\
    GPI-LS & 1.50E+08 \phantom{0}(68\%) & 41 & 150.636 \\
    CapQL & 2.03E+06 \phantom{00}(1\%) & 9 & 7.719 \\
    \midrule
    \end{tabular}
    \caption{Comparison of MORL and EMODPS algorithms}
    \label{tab:results}
\end{table}

\begin{figure}[htb]
    \centering
    \includegraphics[width=\columnwidth]{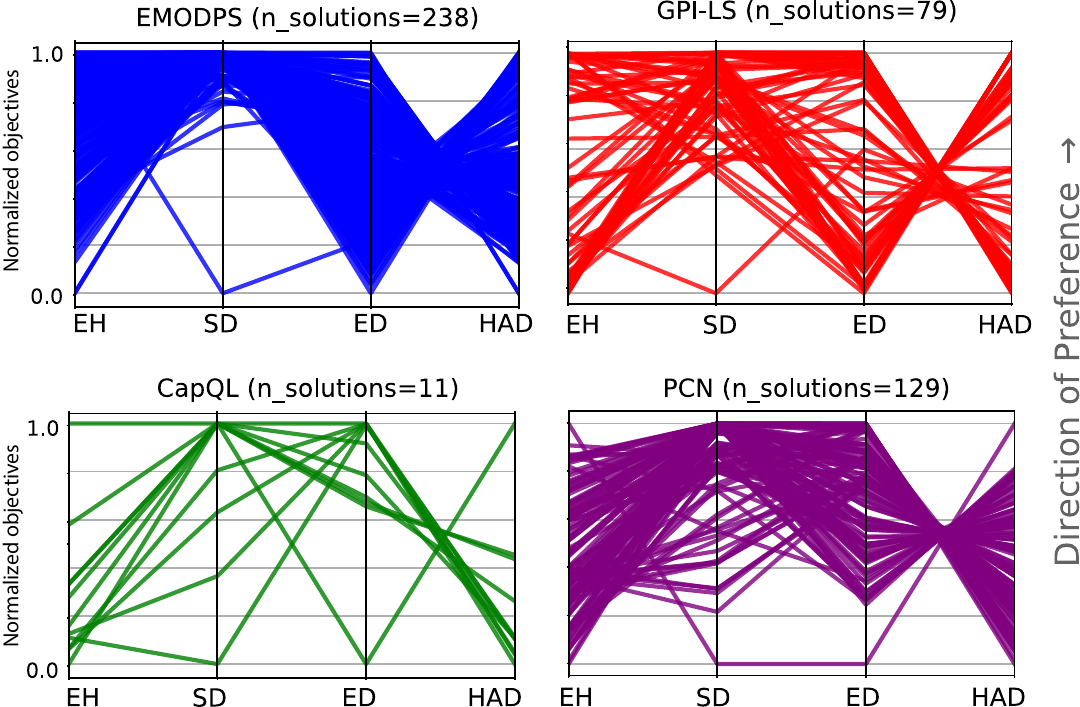} 
    \caption{Parallel coordinate plots with solution sets achieved by each algorithm for the Nile environment.}
    \label{fig:solution_sets}
\end{figure}

Figure~\ref{fig:solution_sets} shows parallel coordinate plots \cite{10.24963/ijcai.2023/755} of each algorithm’s solutions, with the x-axis listing objectives and the y-axis indicating normalized objective values (higher is better); each polyline represents a solution, and an ideal one would be a flat line at 1.0.

The solution sets for each algorithm were generated by merging results from multiple seeds and applying Pareto filtering to retain only non-dominated solutions, with counts shown above each plot. PCN and GPI-LS produce larger merged sets than their average cardinality, suggesting distinct regional coverage across seeds, making exploration an important challenge in this domain. EMODPS achieves a similar size to its mean cardinality, indicating comprehensive exploration within each seed.

\section{Conclusions}

We emphasize the importance of applying MORL algorithms to real-world challenges, as current approaches, often tested on theoretical problems, struggle with practical issues like exploration and scalability. By introducing a complex water management case, we hope to encourage the MORL field to focus on practical challenges when developing new algorithms.  Our results show that current MORL algorithms fall short in comparison to the domain-specific water resource management algorithm, EMODPS. We also highlight two challenges for MORL approaches: scalability and exploration.





\balance

\end{document}